# Optimization of phase-only holograms calculated with scaled diffraction calculation through deep neural networks

Yoshiyuki Ishii, Tomoyoshi Shimobaba*, David Blinder, Tobias Birnbaum, Peter Schelkens, Takashi Kakue, and Tomoyoshi Ito

*1 Graduate School of Engineering, Chiba University, 1-33, Yayoi-cho, Inage-ku, Chiba, 263-8522, Japan*
*2 Department of Electronics and Informatics (ETRO), Vrije Universiteit Brussel, Pleinlaan 2, B-1050 Brussel, Belgium.*
*3 IMEC, Kapeldreef 75, B-3001 Leuven, Belgium*
*\* corresponding author's e-mail: shimobaba@faculty.chiba-u.jp*

Abstract

Computer-generated holograms (CGHs) are used in holographic three-dimensional (3D) displays and holographic projections. The quality of the reconstructed images using phase-only CGHs is degraded because the amplitude of the reconstructed image is difficult to control. Iterative optimization methods such as the Gerchberg-Saxton (GS) algorithm are one option for improving image quality. They optimize CGHs in an iterative fashion to obtain a higher image quality. However, such iterative computation is time consuming, and the improvement in image quality is often stagnant. Recently, deep learning-based hologram computation has been proposed. Deep neural networks directly infer CGHs from input image data. However, it is limited to reconstructing images that are the same size as the hologram. In this study, we use deep learning to optimize phase-only CGHs generated using scaled diffraction computations and the random phase-free method. By combining the random phase-free method with the scaled diffraction computation, it is possible to handle a zoomable reconstructed image larger than the hologram. In comparison to the GS algorithm, the proposed method optimizes both high quality and speed.

1.Introduction

Computer-generated holograms (CGHs) [1-3] are generated by simulating the physical process of holograms on a computer. CGHs are used in holographic 3D displays [4-7] and holographic projections [8-12]. The object light on the CGH plane has a complex amplitude, but the spatial light modulators (SLMs) that display the CGH can only modulate its amplitude or phase. Phase-only SLMs are widely used in practice due to their high light efficiency. The reconstructed image degraded due to a lack of this information. Time division multiplexing [10, 13], down sampling [14, 15], double phase hologram [16-18], and binary amplitude encoding [19] have all been proposed as solutions to this problem. Although these methods provide high-quality reconstructed images, they require an SLM with a high refresh rate and a large spatial bandwidth product.

The Gerchberg-Saxton (GS) algorithm [20-22], one of the optimization methods, can obtain high-quality CGHs by iteratively performing diffraction calculations with known constraints. However, the computing burden of iterative computation is high, and the method tends to fall into a local minimum solution, thus stagnating the improvement of image quality. However, hologram calculation using deep learning has been recently proposed [23-26]. Deep neural networks (DNNs) directly infer CGHs from two-dimensional (2D) and 3D data in these methods. Deep learning algorithms optimize the network parameters of DNNs using a gradient descent algorithm to generate high-quality CGHs during training. After the training is complete, DNNs can directly output CGHs from input data at a high speed. However, in these methods [23-26], the size of reconstructed images is static. Holographic projection [8-12] requires the use of zoomed reconstructed images whose size exceeds CGHs without the need for a zoom lens.

In this paper, we propose a deep learning-based CGH optimization method suitable for holographic projection. This study aims to optimize phase-only CGH (also known as kinoform). In [23-26], DNNs directly generate CGHs from input image data. In contrast, the proposed method generates a CGH from an image by diffraction computation, then inputs it into the proposed DNN, which outputs an optimized CGH. We combine a scaled diffraction calculation [27] with a random phase-free method [28-32] to calculate the input CGHs. This method enables the reconstruction of images larger than the CGH. In general, it is necessary to use random phase when dealing with a reconstructed image larger than the CGH, but highly speckled noise may occur. The proposed method is superior to the GS algorithm in terms of computed speed and image quality because it can obtain reconstructed images with high quality without using random phases. Section 2 describes the proposed method, Section 3 describes the results of simulations and optical experiments, and Section 4 summarizes the study.

## 2. Method

We begin by explaining the GS algorithm, a conventional optimization method, as a comparison method. The GS algorithm applies constraints on the object plane and the CGH plane, respectively, and then iterates to make the reconstructed image of the CGH similar to the original image. When optimizing a phase-only CGH, the amplitude information in the object plane is replaced with the amplitude of the original image while retaining the phase, and then we propagate this new field to the CGH plane. The amplitude of all pixels is unified in the CGH plane, and only the phase information is retained; then we back propagate this new field to the object plane. By repeating this iteration, the reconstructed image gradually approaches the original image closely.

The simulation results of reconstructed images from phase-only CGHs optimized by the GS algorithm are shown in Figure 1. The GS algorithm was iterated 10 times. The original images are given in Figure 1(a), (we used two different images shown in the top and bottom rows), and the reconstructed images without and with the GS optimization are shown in Figures 1(b) and 1(c). The initial phase of the original images was chosen at random. Although the image quality is improved when compared to the unoptimized cases, the computational complexity of the iterative computation is high, and the image quality stagnates, resulting in speckle noise.

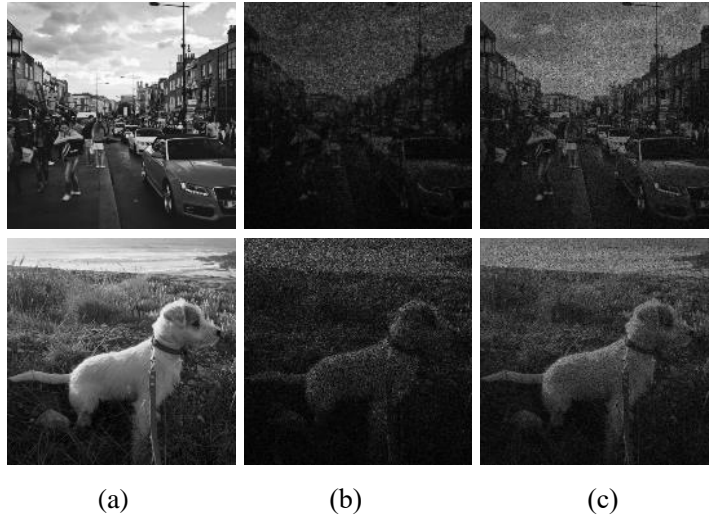

(a) (b) (c)

Fig. 1. Original and reconstructed images: (a) we used two different images as shown in each row, (b) numerically reconstructed images from phase-only CGHs without the GS optimization with random phase [7], and (c) numerically reconstructed images from phase-only holograms optimized by the GS algorithm.

We optimize phase-only CGHs using deep learning to obtain high-quality reconstructed images at high speed. The network used in this study is shown in Figure 2. The network used for training is different from the network used for inference. Phase-only CGHs are computed in the scaled diffraction layer of the network during training. A numerically zoomable CGH can be calculated by changing the sampling pitch set in the source image of the scaled diffraction computation. In this study, we used one of the scaled diffraction computation, Aliasing-Reduced Scaled and Shifted (ARSS) Fresnel diffraction computation [27].

The DNN receives the phase-only CGH generated by the scaled diffraction computation and outputs the optimized phase-only hologram. U-Net [33], one of the network structures, was used as the DNN. The output phase-only CGH is input into the inverse scaled diffraction layer to obtain the reconstructed image. The loss function between the reconstructed image and the original image is computed, and the network parameters of the DNN are optimized using the error backpropagation algorithm to obtain the desired reconstructed image. The trainable parameters are not included in the scaled diffraction layer, and only the DNN parameters are updated.

After training is complete, the two scaled diffraction layers in the network are removed, and a phase-only hologram is generated by a conventional hologram computation algorithm and input into the DNN to obtain the optimal phase-only CGH.

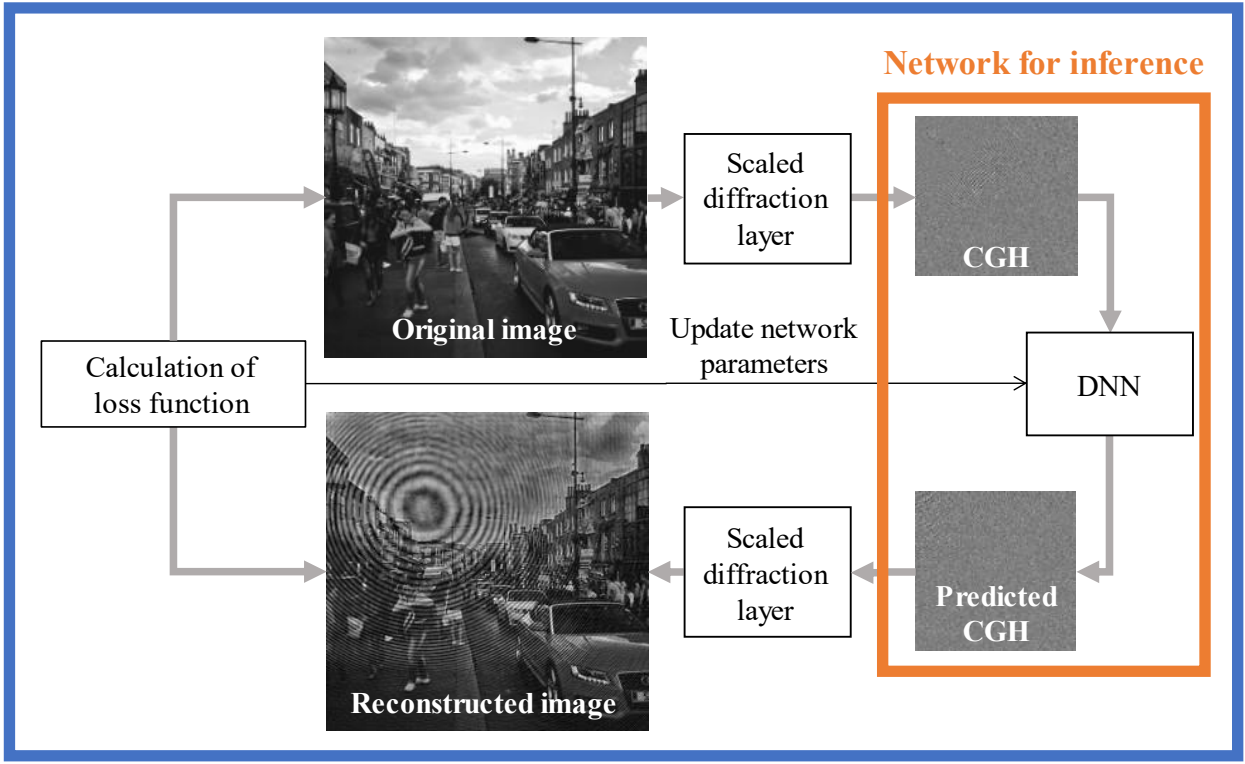

Fig. 2. Proposed method. The structure of the proposed method is different in the training and inferring processes. The network includes the scaled diffraction layers in the training. Phase-only CGHs are input to the DNN after removing the scaled diffraction layer in the inferring.

If the initial phase of the original images is not set in the hologram computation, the object light does not diffuse widely over the CGH plane, and thus an appropriate phase distribution must be given. Random phase is often used as an initial phase distribution, but it causes strong speckle noise in the reconstructed image. In this study, the phase distribution of the virtual convergent light used in the random phase-free method [28] was used as the initial phase. The phase of the virtual convergent light [32] was defined as:

$$w(x_i, y_i) = \exp(-i\pi((x_i + x_{i\text{Offset}})^2 + (y_i + y_{i\text{Offset}})^2)/\lambda f_i), \qquad (1)$$

where $x_i$ and $y_i$ are the coordinates on the object plane, and $x_{i\text{Offset}}$ and $y_{i\text{Offset}}$ are the horizontal and vertical offsets, respectively. The offsets are used to avoid overlap between the non-diffracted light of the SLM and the reconstructed image. $f_i$ is the focal length, and the distance between the CGH and the reconstructed image is $z$. The area of the reconstructed image is $S_i$, and the area of the CGH is $S_h$. We can determine $f_i$ by the following equation

$$f_i: f_i - z = S_i: S_h \times 0.5. \qquad (2)$$

We use scaled diffraction to calculate complex amplitudes $U(x_h, y_h)$ on the CGH plane from an object with the virtual convergent light. A phase-only CGH is calculated by

$$\phi(x_h, y_h) = \arg(U(x_h, y_h)), \qquad (3)$$

where $\arg(\ )$ is the operator taking the argument of the complex number. In this study, we use a bleached phase

CGH [34] as another complex amplitude encoding in addition to the phase-only CGH. Bleached phase CGH is a method to improve the diffraction efficiency of the reconstructed image by bleaching the amplitude hologram [35], which is calculated as

$$\phi(x_h, y_h) = \text{Re}(U(x_h, y_h)). \tag{4}$$

The structure of the U-Net we used is shown in Fig.3. Features are extracted by convolution and max-pooling layers, and a CGH of the same size as the input CGH is obtained by up-sampling layers. Skip connections are added to prevent the loss of detailed information due to max-pooling operations.

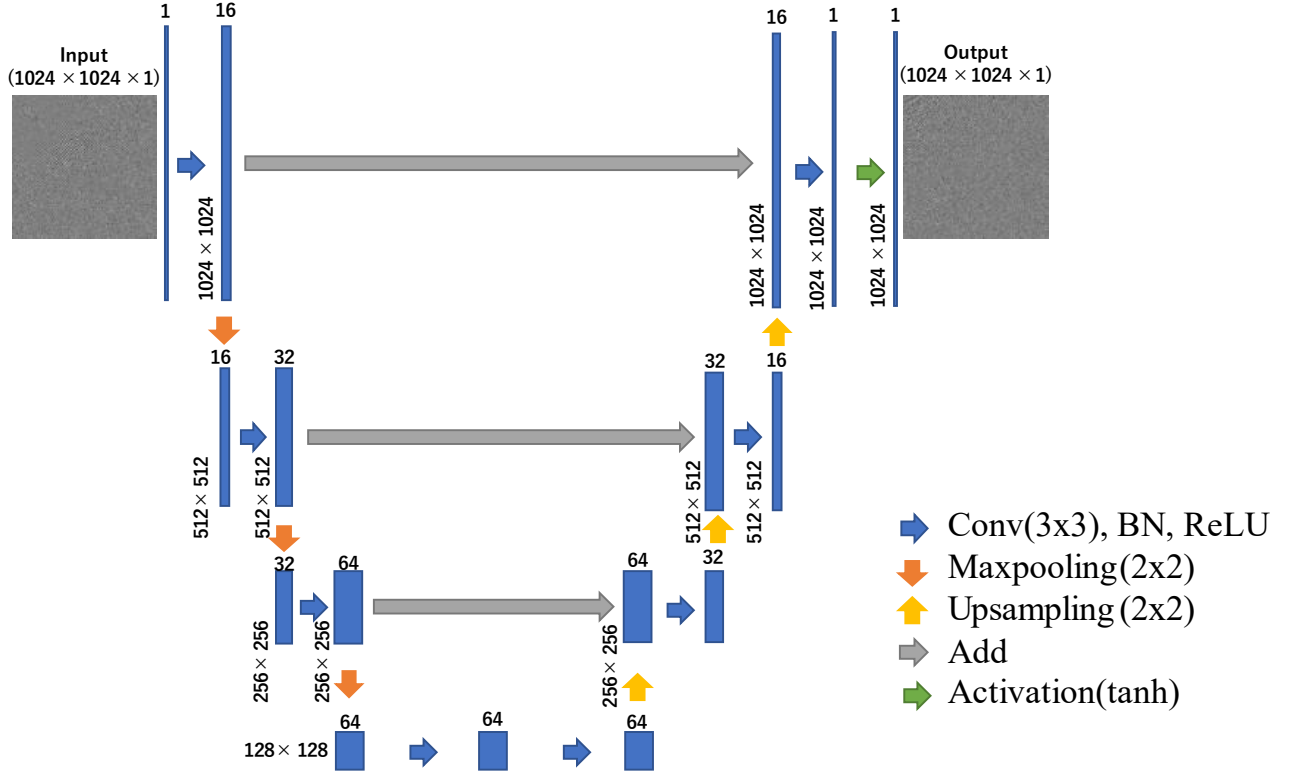

Fig. 3. U-Net-based network structure.

The number at the left of each layer in Fig. 3 shows the number of pixels, which is 128×128 in the lowest layer. The number of channels is shown in the upper part of each layer, and the lowest layer has 64 channels. The size of the convolution filters is all 3×3 with a stride of one, and the convolution layers are zero-pad so that the inputs and outputs of each layer are the same size. We used structural similarity (SSIM) to take the structural information of the reconstructed images into account as the loss function. The parameters of the SSIM were the filter size and standard deviation of 11 and 1.5, $k_1 = 0.01$ and $k_2 = 0.03$, respectively. The optimizer was Adam, and the training batch size was four.

3. Results

Table 1. Calculation parameter.

| Number of pixels [px] | 1024 × 1024 |
|---|---|
| Wavelength [nm] | 532 |
| Pixel pitch of CGH [μm] | 3.74 |
| Pixel pitch of original image [μm] | 18.7 |
| Propagation distance [m] | 0.5 |
| Offset(x-direction) [mm] | 20.48 |
| Offset(y-direction) [mm] | 20.48 |

We used the Open Images Dataset [36, 37] as the training dataset. The image size was resized to 1024 × 1024 px and grayscale. We used 1000 images as the training data, 300 images as the validation data, and 100 images as the test data. The parameters of the CGH calculation are shown in Table 1. By setting the pixel pitch of the original image to five times that of the hologram, the area of a reconstructed image is 25 times larger than that of the CGH. Therefore, the sizes of the CGH and the reconstructed image for simulation and optical experiments are 3.8 mm × 3.8 mm and 19.1 mm × 19.1 mm, respectively.

The proposed method and the GS algorithm were run in a same environment on an Intel Core i7-4790K@4.00GHz with 32GB of RAM and NVIDIA GeForce RTX3070 GPU. We used Python 3.6.13, Keras 2.4.3, and Tensorflow-gpu 2.4.1 as deep learning frameworks.

We input the CGHs computed by the scaled diffraction calculation with the random phase-free method into the trained DNN and reconstruct the inferred CGHs. A comparison of the reconstructed images of the phase-only CGHs is shown in Fig. 4. As a comparison, we also show the results of end-to-end learning, in which the forward diffraction calculation is removed from the network in Fig. 2 and the holograms are directly inferred from the original images (hereafter referred to as "end-to-end learning"). From left to right, we show the original image, non-optimization (we only used a random phase-free method), GS algorithm (10 iteration), the proposed method, and end-to-end learning.

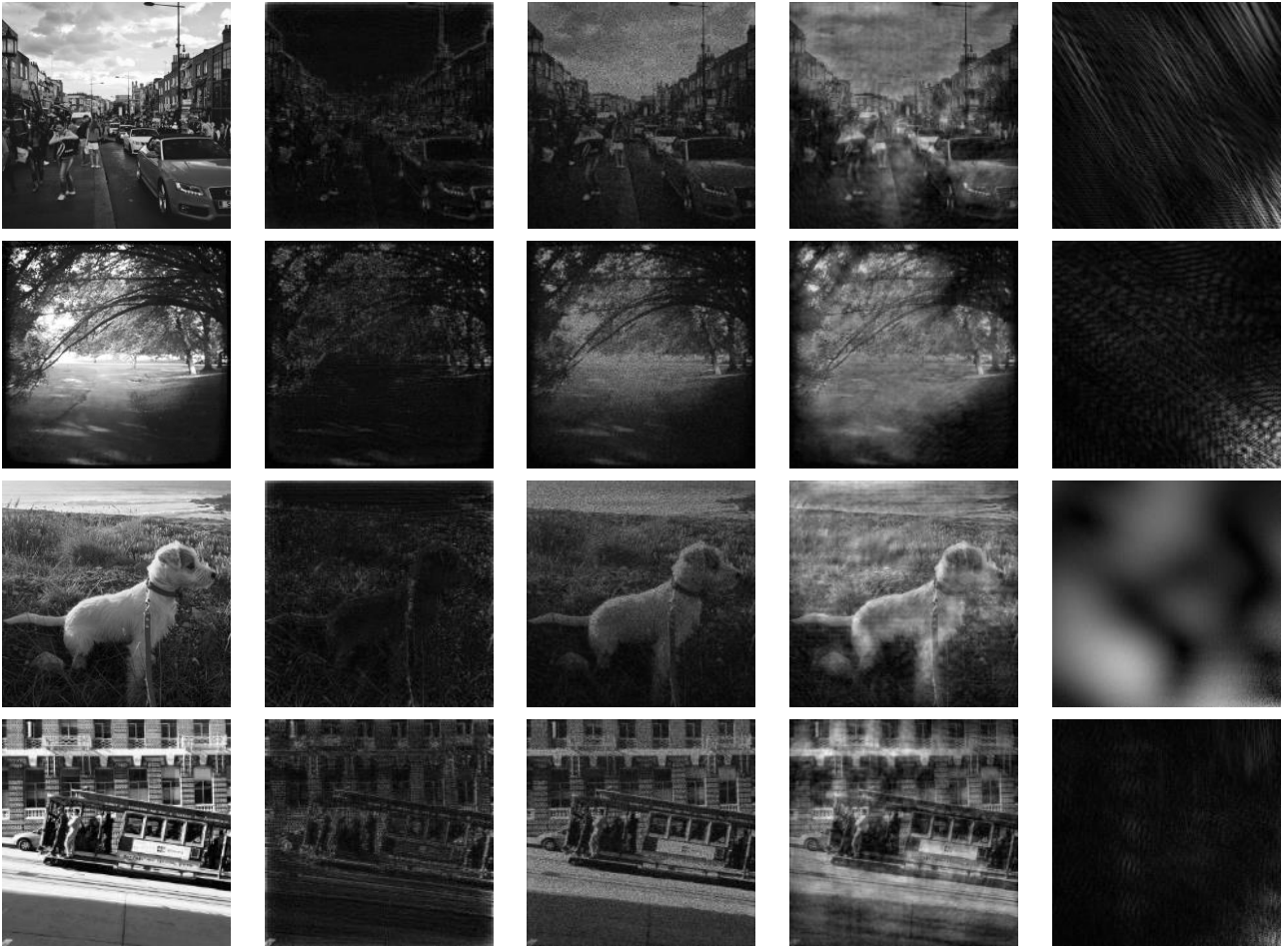

Fig. 4. Comparison of the reconstructed images of phase holograms From left to right: original image, non-optimization (random phase-free method), GS algorithm, proposed method, and end-to-end learning.

The random phase-free method resulted in an edge-enhanced reconstructed image. This is a common phenomenon in phase-only holograms [38]. The end-to-end learning failed to generate correct CGHs. Although the proposed method can be more efficient than the GS algorithm and is capable of reconstructing details of the images, the images are entirely superimposed with noise.

Following that, we used a bleached phase CGH [34] to perform the same training and inference as in Fig. 4. A comparison of the reconstructed images is shown in Fig. 5. The original image, non-optimization (random phase-free method), GS algorithm (10 iteration), the proposed method, and end-to-end learning are shown from left to right.

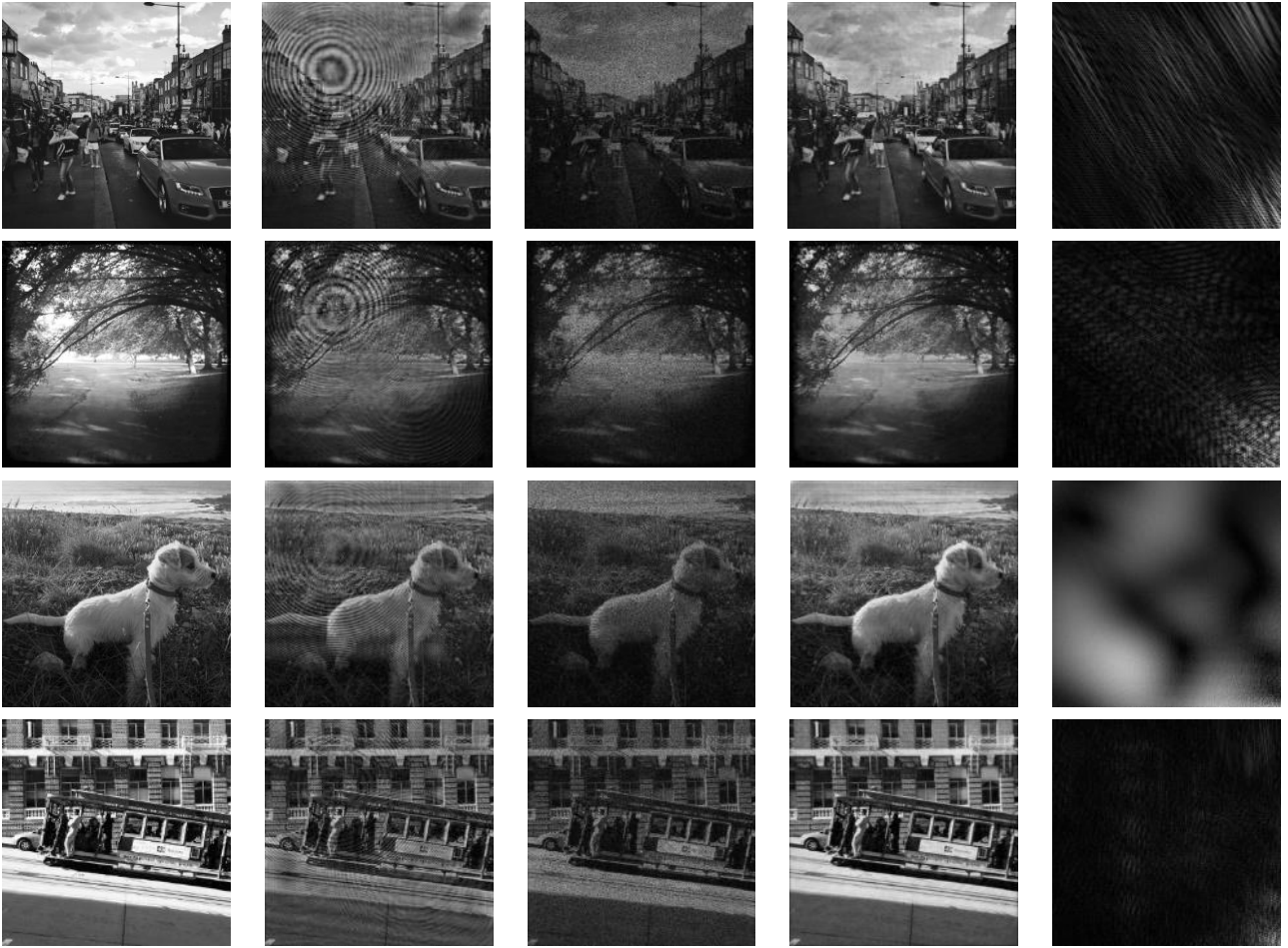

Fig. 5. Comparison of reconstructed images of bleached phase hologram. From left to right: original image, non-optimization (random phase-free method), GS algorithm, proposed method, and end-to-end learning.

The proposed method generates images similar to the original images when utilizing bleached phase CGH, as shown in Fig. 5. The proposed method is able to optimize holograms for high image quality, unlike the GS algorithm. The CGHs inferred by end-to-end learning do not generate the desired reconstructed images, as shown in Figs. 4 and 5. This is because the image domains of the original image and the CGHs are significantly different, making it difficult to train the DNN [39].

We compare the conventional methods of the random phase-free method and the GS algorithm with the proposed method using quantitative image quality assessment. The average values of peak signal-to-noise ratio (PSNR) and SSIM for 100 test data are shown in Table 2. In the evaluation, all images were converted to 8-bit grayscale images.

Table 2. Quantitative assessment of reconstructed images.

| | | PSNR [dB] | SSIM |
|---|---|---|---|
| Phase-only CGH | Non-optimization (Random phase-free method) | 7.72 | 0.274 |
| | GS algorithm (10 iteration) | 12.91 | 0.309 |
| | Proposed method | 14.07 | 0.614 |
| Bleached phase CGH | Non-optimization (Random phase-free method) | 14.99 | 0.544 |
| | GS algorithm (10 iteration) | 13.13 | 0.315 |
| | Proposed method | 21.73 | 0.822 |

Both the phase-only CGHs and bleached phase CGHs show better results in terms of PSNR and SSIM than the non-optimization method and the GS algorithm, as shown in Table 2. In particular, the image quality assessment is better than the other methods when the proposed method optimizes the bleached phase of CGH. In addition, the GS algorithm with 10 iterations takes 60 ms to optimize a CGH, whereas the proposed method can infer a CGH in 27 ms, which is faster than the GS algorithm.

We show results of optical experiments. The optical setup and the optical reconstructions of bleached phase CGHs that showed better results in the simulation results are shown in Figs. 6 and 7. We used a green laser at 532 nm and Holoeye GAEA2 as the SLM. The optical reconstructions were directly captured on the image sensor.

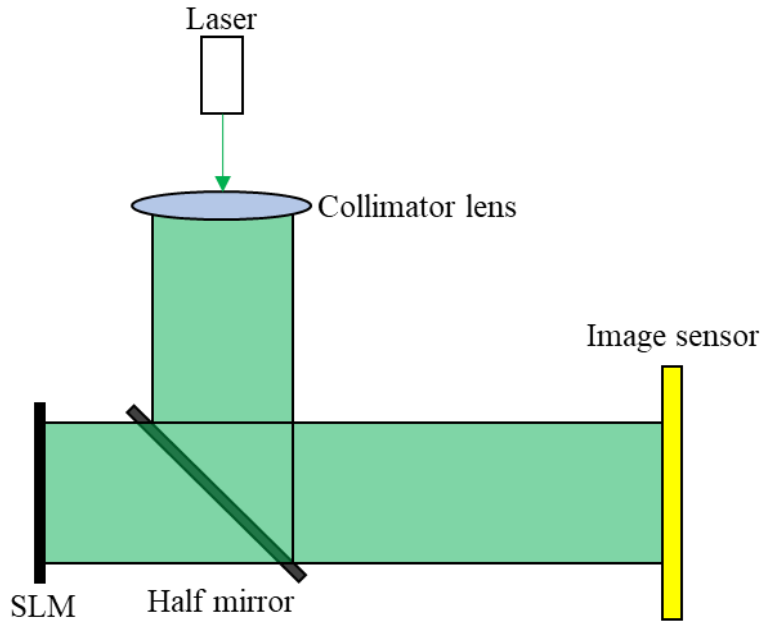

Fig. 6. Optical system.

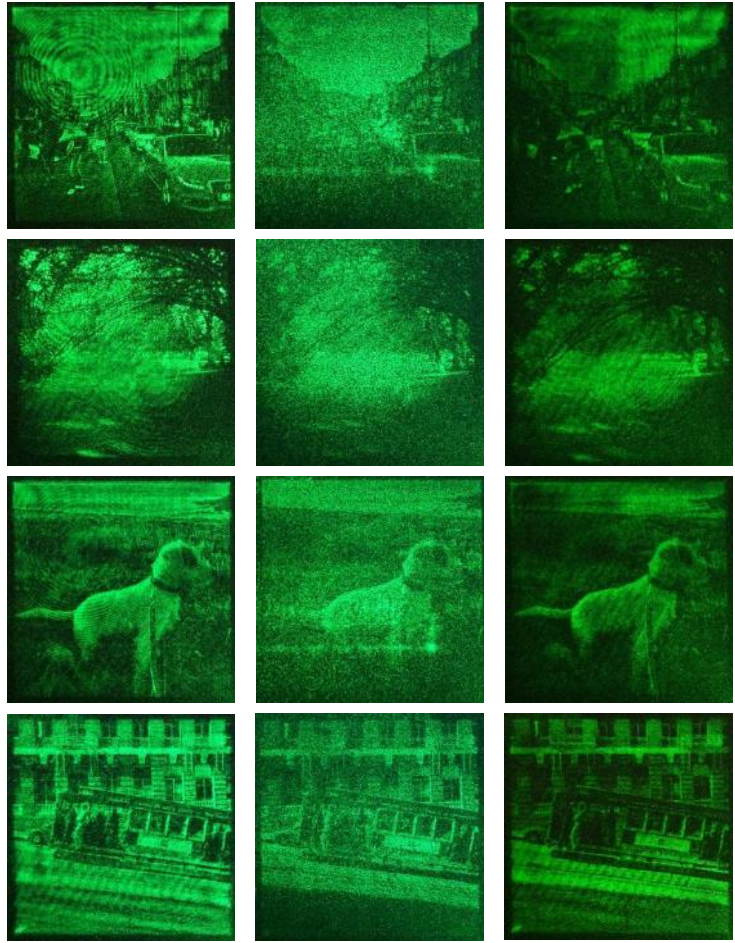

Fig. 7. Optical reconstructed images of the bleached phase holograms. From left to right, non-optimization, GS algorithm (10 iteration), and the proposed method.

In Fig. 7, the GS algorithm does not represent fine details because of speckle noises, and the non-optimization method includes spherical phase errors introduced by the random phase-free method, but the proposed method improves the overall image quality, including the fine details. The reconstructed images at the bottom row of Fig. 7 is magnified in Fig. 8. The spherical phase noises are observed in the non-optimization method as shown in Fig. 5, but in the proposed method; the spherical phase noises are removed and the image quality is improved.

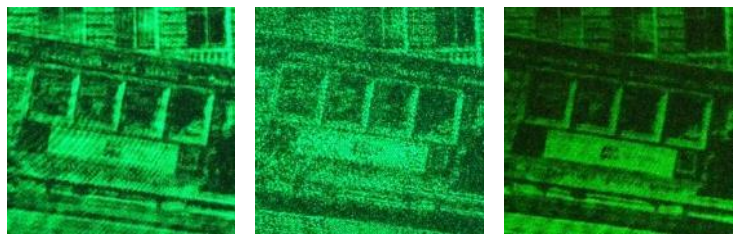

Fig. 8. Enlarged images of the optical reconstructed images of the bleached phase of CGH. From left to right: non-optimization, the GS algorithm (10 iterations), and the proposed method.

Finally, we show a zoomable holographic projection. After training the proposed DNN with a constant pixel pitch five times larger than that of the CGH, we generated CGHs by changing the pixel pitch of the original image from twice to five times the CGH. The generated CGHs were then input into the DNN without changing the weight parameters of the DNN, yielding optimized CGHs as an output. A snapshot of the simulated and optically reconstructed movie while changing the pixel pitches on the original image is shown in Figure 9. The movie was taken in which the reconstructed images zoomed in and out by displaying the optimized CGHs sequentially.

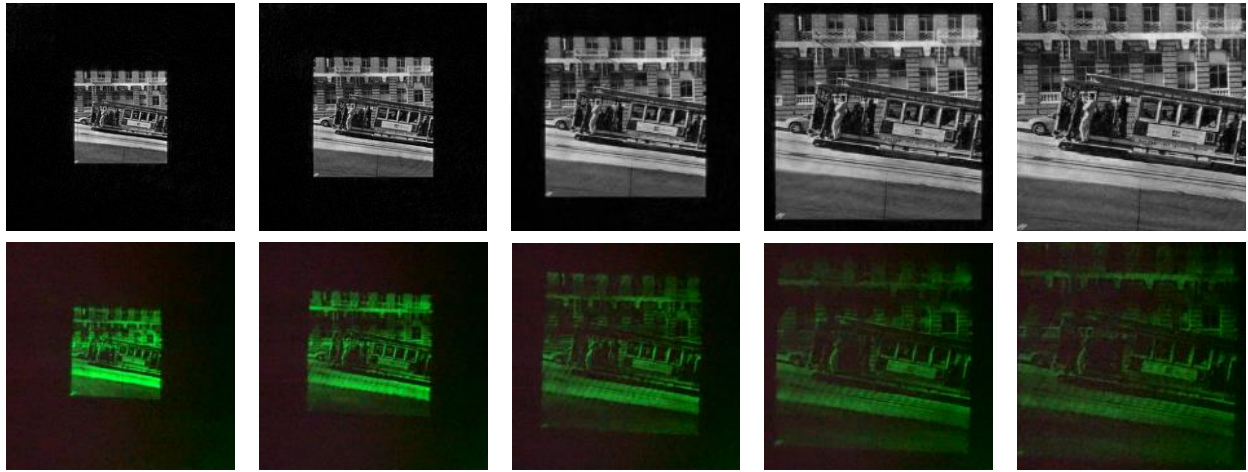

Fig. 9. Snapshots of a movie in which the size of the reconstructed image is enlarged and reduced by changing the pixel pitch of the original image through computation. The pixel pitch of the reconstructed image is two to five times larger than that of the hologram. The top and bottom rows show the simulation and optical experiment, respectively.

4. Conclusion

By optimizing the hologram, we obtained reconstructed images larger than CGHs by combining deep learning, the random phase-free method, and ARSS Fresnel diffraction computation. The simulation and optical experiments showed that the proposed method outperforms the GS algorithm and the random phase-free algorithm in terms of image quality and computational speed.

Acknowledgements

This work was partially supported by Japan Society for the Promotion of Science (JSPS) KAKENHI Grant Numbers 19H04132 and 19H1097, the joint JSPS-FWO scientific cooperation program (VS07820N) and the Research Foundation - Flanders (FWO), Junior postdoctoral fellowship (12ZQ220N).